\documentclass{article}

\usepackage{arxiv}
\usepackage[utf8]{inputenc} 
\usepackage[T1]{fontenc}    
\usepackage{hyperref}       
\usepackage{url}            
\usepackage{booktabs}       
\usepackage{amsfonts}       
\usepackage{nicefrac}       
\usepackage{microtype}      
\usepackage{lipsum}		
\usepackage{graphicx}
\usepackage{natbib}
\usepackage{doi}

\usepackage{algorithm}
\usepackage{algpseudocode}
\usepackage{multirow}
\usepackage{xcolor}
\usepackage[most]{tcolorbox}
\usepackage{array}
\usepackage{cancel}
\usepackage{comment}

\definecolor{keywordcolor}{RGB}{0,0,139}
\definecolor{variablecolor}{RGB}{0,100,0}
\definecolor{green}{RGB}{0,255,0}
\definecolor{blue}{RGB}{0,0,255}
\definecolor{orange}{RGB}{255,165,0}
\definecolor{red}{RGB}{255,0,0}
\definecolor{purple}{RGB}{128,0,128}
\definecolor{cyan}{RGB}{0,255,255}
\definecolor{magenta}{RGB}{255,0,255}
\definecolor{yellow}{RGB}{255,255,0}
\definecolor{brown}{RGB}{139,69,19}
\definecolor{gray}{RGB}{128,128,128}
\definecolor{pink}{RGB}{255,182,193}
\definecolor{teal}{RGB}{0,128,128}
\definecolor{olive}{RGB}{128,128,0}
\definecolor{lightblue}{RGB}{173,216,230}
\definecolor{darkblue}{RGB}{0,0,139}
\definecolor{userpromptcolor}{RGB}{148,255,255}

\newtcolorbox{mybox}[1]{
  enhanced,
  colback=lightblue!30,
  colframe=black,
  arc=4mm,
  fonttitle=\color{white}\bfseries,
  title=#1,
  coltitle=white,
  boxrule=0.5mm
}

\newtcolorbox{userpromptbox}[1]{
  enhanced,
  colback=userpromptcolor!30,
  colframe=black,
  arc=4mm,
  fonttitle=\color{white}\bfseries,
  title=#1,
  coltitle=white,
  boxrule=0.5mm
}

\title{Enhancing Computer Programming Education with LLMs: A Study on Effective Prompt Engineering for Python Code Generation}

\author{%
    \href{https://orcid.org/0000-0002-9244-8798}{\includegraphics[scale=0.06]{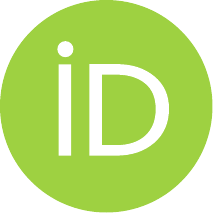}\hspace{1mm}Tianyu Wang}\thanks{These authors contributed equally to this work.}\\
    Mercy University\\
    Math \& Computer Science Department\\
    555 Broadway,\\
    Dobbs Ferry, NY 10522, USA\\
    \texttt{twang4@mercy.edu}
    \and
    \href{https://orcid.org/0000-0002-3473-6097}{\includegraphics[scale=0.06]{orcid.pdf}\hspace{1mm}Nianjun Zhou}\footnotemark[1]\\
    IBM Research\\
    1101 Kitchawan Road, Route 134\\
    Yorktown Heights, NY 10598, USA\\
    \texttt{jzhou@us.ibm.com}
    \and
    \href{https://orcid.org/0000-0002-9874-6972}{\includegraphics[scale=0.06]{orcid.pdf}\hspace{1mm}Zhixiong Chen}\\
    Mercy University\\
    Math \& Computer Science Department\\
    555 Broadway,\\
    Dobbs Ferry, NY 10522, USA\\
    \texttt{zchen@mercy.edu}
}

\renewcommand{\shorttitle}{Enhancing Computer Programming Education with LLMs}

\hypersetup{
pdftitle={Enhancing Computer Programming Education with LLMs: A Study on Effective Prompt Engineering for Python Code Generation},
pdfsubject={Prompt Engineering of LLM in Education},
pdfauthor={Tianyu Wang, Nianjun Zhou, Zhixiong Chen},
pdfkeywords={Prompt Engineering, Large Language Models (LLMs), Computer Programming, Code Quality Evaluation,
Education},
}

\begin{document}
\maketitle

\begin{abstract}

Large language models (LLMs) and prompt engineering hold significant potential for advancing computer programming education through personalized instruction. This paper explores this potential by investigating three critical research questions: the systematic categorization of prompt engineering strategies tailored to diverse educational needs, the empowerment of LLMs to solve complex problems beyond their inherent capabilities, and the establishment of a robust framework for evaluating and implementing these strategies. Our methodology involves categorizing programming questions based on educational requirements, applying various prompt engineering strategies, and assessing the effectiveness of LLM-generated responses. Experiments with GPT-4, GPT-4o, Llama3-8b, and Mixtral-8x7b models on datasets such as LeetCode and USACO reveal that GPT-4o consistently outperforms others, particularly with the "multi-step" prompt strategy. The results show that tailored prompt strategies significantly enhance LLM performance, with specific strategies recommended for foundational learning, competition preparation, and advanced problem-solving. This study underscores the crucial role of prompt engineering in maximizing the educational benefits of LLMs. By systematically categorizing and testing these strategies, we provide a comprehensive framework for both educators and students to optimize LLM-based learning experiences. Future research should focus on refining these strategies and addressing current LLM limitations to further enhance educational outcomes in computer programming instruction.


\end{abstract}


\section{Introduction}\label{section:introduction}

\begin{figure*}[ht]
\centering
\includegraphics[width=8cm]{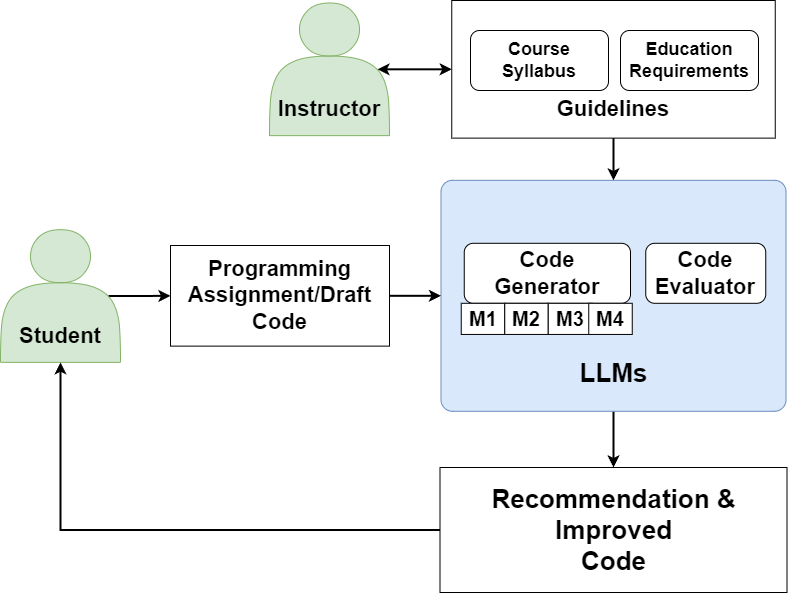}
\caption{Flowchart illustrating the users in code generation and evaluation process}
\label{fig:system}
\end{figure*}

In the rapidly evolving field of Artificial Intelligence (AI), Large Language Models (LLMs) \citep{bommasani2021opportunities, touvron2023llama}, a breakthrough in AI, have shown remarkable potential in a variety of applications, including natural language processing, content creation, and more recently, in code generation. Numerous LLM models, such as ChatGPT and LLaM, have shown their capability to cater to a variety of task domains, as Fig \ref{fig:system} shown, ranging from question answering to the generation of code snippets,  \citep{radford2019language, bommasani2021opportunities, ouyang2022training, liu2023pre, touvron2023llama}. Furthermore, in the evolving landscape of computer science education, integrating advanced technologies has become increasingly pivotal. Many Integrated Development Environments (IDEs) have equipped with LLMs to generate code in software development tools, such as VS Code\footnote{https://code.visualstudio.com/}. As the demand for proficient programmers grows, so does the necessity for innovative and effective teaching methods. These usages of tools draw attention to developing software vulnerabilities and security concerns \citep{asare2023github, pearce2022asleep, dakhel2023github}.

This paper presents significant contributions to the field of AI-assisted programming education by focusing on the optimization of Python code generation with LLMs for educational applications. Our research addresses critical aspects of how LLMs can be effectively utilized to create personalized and adaptive learning environments that cater to diverse educational needs. We systematically investigate and categorize prompt engineering strategies, tailoring them to specific educational objectives and problem types. By doing so, we enable educators and students to leverage LLMs in a way that maximizes their problem-solving potential and instructional effectiveness.

Our contributions are threefold:
\begin{enumerate}
    \item \textbf{Categorization of prompt engineering strategies}. We develop a comprehensive categorization of prompt engineering strategies that align with various educational requirements, ranging from foundational knowledge and skills to competition-level challenges and advanced problem-solving tasks. This categorization provides a structured approach for educators to customize prompts and optimize learning pathways for their students.
    \item \textbf{Explore prompt engineering impact on LLM performance in code generation}. We propose a robust framework designed to explore and validate various prompt engineering strategies, assessing their impact on LLM performance in generating Python code. Our findings underscore the effectiveness of different prompt strategies in guiding students through complex problem-solving processes, thereby enhancing the role of LLMs as facilitators in the educational journey.
    \item \textbf{Provide a general prompt guidelines for different educational purpose}. Following the evaluation of prompt engineering strategies using various datasets, we present comprehensive guidelines for educators. These guidelines facilitate the systematic use of LLMs to generate superior code and optimize LLM-based learning experiences. By ensuring the efficient application of diverse prompt strategies, these guidelines aim to enhance educational outcomes in computer programming instruction across different educational requirements.
\end{enumerate}

By advancing the application of LLMs in Python code generation for educational purposes, our research contributes to the broader goal of integrating AI technologies into educational practices, paving the way for more dynamic, personalized, and effective learning experiences in the field of computer programming.

\section{Related Work}\label{section:related_work}

Recent research in the educational sector has prominently featured the application of Large Language Models (LLMs) to enhance learning outcomes, particularly in the programming domain \cite{denny2023computing}. This review synthesizes findings from key studies that illustrate the diverse roles LLMs play in education, from interactive assistance in computer science courses to the evaluation of programming skills. \citet{murr2023testing} focused on the effectiveness of LLMs in generating code, emphasizing the critical role of prompt specificity. Many studies have demonstrated the efficacy of integrating AI code generators in introductory programming courses, as evidenced by research conducted by \cite{finnie2022robots}, \cite{hellas2023exploring}, and \cite{kazemitabaar2023studying}. In addition, \cite{kiesler2023large} assessed the capabilities of ChatGPT-3.5 and GPT-4 in solving introductory Python programming tasks sourced from CodingBat. \cite{pearce2022pop} explored the application of LLMs in reverse engineering tasks and exhibited promising results. Supporting discussions on LLMs' application in programming education, particularly with development assistant, additional references such as \cite{asare2023github}, \cite{pearce2022asleep}, \cite{dakhel2023github} and \cite{denny2023chat} provide insights into the integration of AI tools within software development environments. These studies collectively underscore the transformative impact of LLMs on programming education, suggesting avenues for future research in optimizing their use for educational enhancement and addressing broader software development challenges \cite{chen2021evaluating}.

While LLMs were impressively successful in generating code for different purpose in education and production, they stumbled when confronting real-world security and risks concerns.
\cite{bommasani2021opportunities} examine the potential and risks associated with LLMs and highlights their content creation capabilities and warns about potential issues like bias, misinformation, and homogenization. ~\cite{dakhel2023effective} leveraged LLMs for generating unit tests in software development. To overcome the inaccurate response from LLM, a new discipline or guideline called 'Prompt Engineering', which includes specific strategies for maximizing the capability of LLM, applies to us with modification \citep{reynolds2021prompt, liu2023pre}. delve into methodologies for leveraging the inherent capabilities of narratives and cultural anchors to intricately encode intentions and strategies, thereby facilitating a structured breakdown of problems into their constituent elements prior to reaching conclusions \cite{reynolds2021prompt}. Expanding upon this notion, introduce a novel approach termed least-to-most prompting, which systematically deconstructs complex issues into manageable sub-problems, addressing them sequentially to enhance problem-solving efficiency in LLMs \cite{zhou2022least}. Complementing these insights, demonstrate the natural emergence of reasoning capabilities within sizable LLMs through what is known as chain-of-thought prompting. This technique involves the use of select demonstrations that guide the model through a thought process, thereby facilitating the comprehension and solving of tasks \cite{wei2022chain}. In a similar vein, propose an innovative "Ask Me Anything" (AMA) prompting strategy that iteratively employs the LLM itself to reformulate task inputs into a more effective question-and-answer format, thereby significantly augmenting the performance of LLMs \cite{arora2022ask}. Additionally, explore the potential of refining language models' task execution and instruction-following capabilities through the integration of human feedback, marking a significant step towards more interactive and adaptive LLMs \cite{ouyang2022training}. The study by \cite{white2023prompt} on prompt pattern catalog to enhance prompt engineering with ChatGPT provides a comprehensive overview of best practices and patterns in prompt engineering, highlighting its importance in optimizing LLM outputs for specific tasks.


The code quality generated by LLMs holds paramount importance in applications spanning educational contexts and real-world production environments and many code evaluation studies have been introduced. \cite{hendrycks2021measuring} unveiled APPS, a benchmark specifically designed for code generation tasks. This benchmark assesses the capability of models to interpret arbitrary natural language specifications and produce Python code that meets the specified requirements. Furthermore, \cite{chen2021evaluating} introduced HumanEval, an innovative evaluation set aimed at measuring the functional correctness of programs synthesized from docstrings. \cite{xu2022systematic} conducted a comprehensive assessment of the largest code-generating models available, spanning multiple programming languages. They introduced a novel model, PolyCoder, which demonstrated superior performance in generating C programming code, outperforming its counterparts. \cite{liu2023your} developed EvalPlus, a comprehensive framework for the evaluation of code synthesis. This framework is meticulously designed to benchmark the functional correctness of code generated by LLMs with a high degree of rigor. \cite{white2023chatgpt} presented a more systematic methodology for the cataloging of software engineering patterns. This study classifies various patterns and delves into numerous prompt strategies that have been employed to enhance code quality and system design. In a comparative study, \cite{murr2023testing} analyzed the efficacy of code produced by different LLMs across 104 customized Python challenges, utilizing widely recognized metrics such as the pass rate for assessment.


Beyond conventional methodologies, deep learning techniques have increasingly been applied to the evaluation of code. \cite{kanade2020learning} explored the capabilities of a finely tuned CuBERT model, revealing that it surpasses traditional methods in source code evaluation. This advantage was observed even with limited training and a smaller number of labeled examples. \cite{ciniselli2021empirical} presented an empirical study employing a RoBERTa model to assess its efficiency in code completion tasks from various angles. The findings indicate that BERT-based models are a promising avenue for enhancing code completion capabilities. \cite{wang2023analyzing} proposed an approach for the automatic assessment of code quality using a BERT model that has been meticulously fine-tuned with specific datasets, offering a groundbreaking perspective on the evaluation of code quality.




\section{Problem Statement}
\label{section:problem_statement}







In computer programming education, LLMs and prompt engineering hold promise for personalized instruction. By harnessing LLMs' capability to comprehend and respond to natural language prompts, we can create adaptive learning environments that dynamically adjust to a education requirement and problem-solving approach. However, a pivotal question persists:\textbf{ can we fully exploit this potential to enhance educational outcomes?}

This paper investigates this very issue by focusing on three key research questions:

\begin{enumerate}
    \item \textbf{Can we systematically categorize prompt engineering strategies tailored to various educational requirements and question types?} This categorization would enable educators and students to customize prompts according to specific problem types and educational objectives, thereby maximizing the effectiveness of computer programming instruction. Understanding these categories is crucial for developing targeted and efficient learning pathways.
    \item \textbf{Do different strategies empower LLMs to address problems beyond their immediate solution capabilities?} This exploration aims to unlock the potential for LLMs to guide students through complex or unfamiliar problem-solving processes. By identifying strategies that extend LLM capabilities, we can enhance their role as facilitators in the learning journey, providing support even in challenging scenarios.
    \item \textbf{Can we establish a robust framework for testing the effectiveness of various prompt engineering strategies and provide comprehensive guidelines for their implementation?} This framework would allow educators to systematically evaluate and optimize LLM-based learning experiences, ensuring maximum benefit for students. By developing and validating these guidelines, we can offer a structured approach to enhance educational practices and outcomes through LLMs.
\end{enumerate}
By addressing these questions, this research aims to pave the way for a future of personalized and adaptive coding education, empowering students to approach problems with confidence and a deeper understanding of core programming concepts.
\section{Methodology}
\label{section:methodology}




\begin{figure*}[ht]
\centering
\includegraphics[width=16cm]{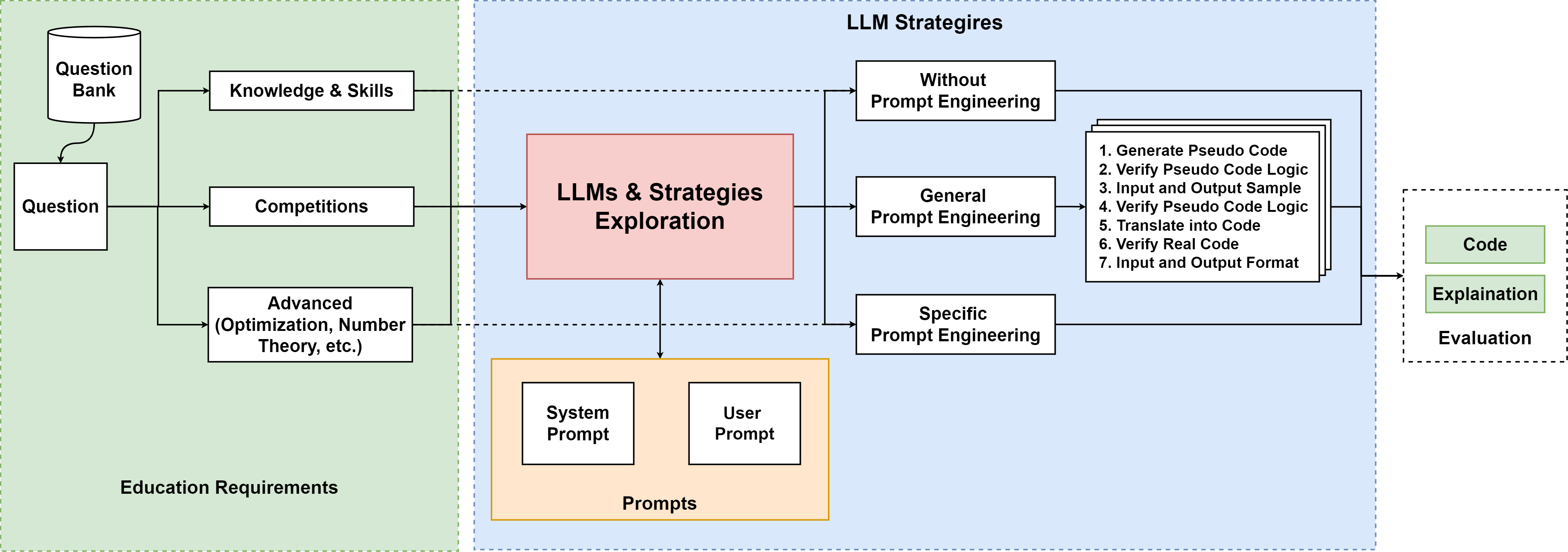}
\caption{Conceptual Diagram Highlighting the Interaction Between LLMs and Prompt Engineering}
\label{fig:model}
\end{figure*}

To address our research problem, we designed a comprehensive model, as illustrated in Fig \ref{fig:model}. This model outlines the process of categorizing questions, applying various prompt engineering strategies, and evaluating the effectiveness of LLM-generated responses. The methodology is structured into three primary steps:

\begin{enumerate}
\item \textbf{Categorization of Questions Based on Educational Requirements}. The first step involves categorizing computer programming questions according to different educational requirements. This categorization helps in tailoring prompt engineering strategies to the specific needs of learners. We classify the questions into three distinct levels:
    \begin{itemize}
        \item \textbf{Knowledge and Skills}: This category focuses on fundamental algorithms and data structures, preparing students for coding interviews. It aims to build a solid foundation in programming by familiarizing students with essential concepts and techniques.
        \item \textbf{Competitions}: This category is designed to enhance programming skills for higher-level competitions. It includes problems that challenge students to think critically and innovatively, helping them gain recognition that can benefit their academic and professional careers.
        \item \textbf{Advanced Complex Problems}: This category addresses advanced topics such as algorithm design, optimization, and number theory. It is intended for students who have a strong grasp of basic concepts and are looking to tackle more sophisticated and intricate problems.
    \end{itemize}

\item \textbf{Exploration of Prompt Engineering Strategies}. In the next step, we explore different prompt engineering strategies on the categorized questions. There are three major prompt engineering strategies that we employ:
    \begin{itemize}
        \item \textbf{Without Prompt Engineering}: In this approach, we present the original question to the LLM without any additional guidance. This strategy tests the LLM's inherent ability to comprehend and solve problems without external aid.
        \item \textbf{General Prompt Engineering}: This method involves providing some general prompt templates to help the LLM understand the problem better. For example, we include the requirements of question constraints, test cases, and guidelines to assist the LLM in breaking down the problem into logical sub-problems. This approach serves as a framework to improve the LLM's problem-solving capabilities.
        \item \textbf{Specific Prompt Engineering}: This strategy provides detailed instructions on how to address the problem, rather than using a generic approach. For instance, we may instruct the LLM to consider extreme scenarios or treat the problem as an optimization task. This method is particularly useful for challenging and complex problems that require specialized solutions.
    \end{itemize}

\item \textbf{Evaluation of LLM Responses}. After applying the appropriate prompt engineering strategies, we evaluate the LLM-generated responses using different metrics. The evaluation process involves:

    \begin{itemize}
        \item \textbf{Correctness}: Assessing the correctness of the generated code. We verify if the code meets the problem requirements and constraints.
        \item \textbf{Effectiveness}: Measuring the overall effectiveness of the LLM's code. This includes testing the code against efficiency in terms of time and memory complexity. More details about evaluation can be found at Section \ref{section:evaluation}.
    \end{itemize}

\end{enumerate}

By following this structured methodology, we aim to systematically investigate the potential of LLMs and prompt engineering in enhancing computer programming education. The insights gained from this research will provide valuable guidelines for educators to optimize LLM-based learning experiences for maximum student benefit.

\subsection{General Prompt Engineering}

\begin{table*}
\centering
\caption{Prompt Configurations for Model}
\begin{tabular}{|l|l|l|}
\hline
\textbf{Model} & \textbf{System Prompt} & \textbf{User Prompts and Chains} \\
\hline
base & \multirow{3}{1.5cm}{See Appendix~\ref{section:prompt}} & Question | llm -> Code \\
\cline{1-1}\cline{3-3}
example (1-shot) & & Question | llm -> Code | Example -> Code\\
\cline{1-1}\cline{3-3}
dynamic example & & Question | llm -> Dynamic Related Example -> Code\\
\cline{1-1}\cline{3-3}
guide & & Question | llm -> Code | General Guide\footnote{See appendix~\ref{section:general_guide_content}} -> Code\\
 \cline{1-1}\cline{3-3}
multi & & Question | llm -> multi-step chat guide | llm -> Code\\
\cline{1-1}\cline{3-3}
all-in-one & & Question | All-in-one | llm -> Code \\
\cline{1-1}\cline{3-3}
\hline
\end{tabular}
\label{tab:prompt}
\\\vspace{2mm} 
 \raggedright
\end{table*}
Prompt engineering is crucial in the development and optimization of LLMs due to its significant impact on the performance and accuracy of these models. The choice of prompts directly influences the output generated by LLMs, as they determine how the model interprets and responds to the given input. By experimenting with and testing different prompts for the same question, researchers can identify the most effective phrasing and structure, thereby enhancing the model's ability to generate relevant, coherent, and contextually appropriate responses. This iterative process of prompt refinement ensures that the LLMs are not only robust but also adaptable to a wide range of applications, thereby maximizing their utility in various domains. As Table \ref{tab:prompt} shown, to explore and compare different prompts in LLMs, we use the following prompt settings in our study. 

\begin{itemize}
    \item \textbf{Prompt 1 (\texttt{base}):} This configuration serves as our baseline. In this setup, the model is presented with the fundamental structure of a problem statement. The objective is to assess the model's inherent capability to understand and generate responses with minimal guidance.
    \item \textbf{Prompt 2 (\texttt{example (1-shot)}):} In this prompt, the model is initially provided with the basic structure of a problem statement. Subsequently, we introduce a single of high-quality code example to the LLM. This configuration aims to evaluate the impact of enhanced prompt structures on the model's output, particularly focusing on how the inclusion of an example influences the generated responses.
    \item \textbf{Prompt 3 (\texttt{dynamic example}):} Similar to prompt 2, this prompt initially presents the problem statement. However, instead of using a static, high-quality code example, the LLM is tasked with generating a related but distinct example. The goal here is to assess the model's adaptability and ability to produce a code solution based on dynamically generated examples.
    \item \textbf{Prompt 4 (\texttt{guide}):} This prompt incorporates general coding guidelines alongside the problem statement. The intention is to guide the LLM towards generating code that adheres to broad quality standards. Details regarding the content of these general guidelines are provided in Appendix~\ref{section:general_guide_content}.
    \item \textbf{Prompt 5 (\texttt{multi}):} This configuration employs a multi-step conversational prompt strategy with LLMs. Initially, it provides a multiple turns of interaction, where the model responds to ongoing inputs that build on previous context. Instructors or students evaluate each suggestion, providing feedback to guide the LLM towards optimal solutions. This approach usually suits problems requiring sustained engagement like tutoring or solving complex problems. Specifically, 
        \begin{enumerate}
            \item \textbf{Generate Pseudo Code:} The initial step involves generating pseudo code based on the question. This pseudo code acts as an intermediate representation of the solution, outlining the logical steps necessary to solve the problem without delving into specific syntax. The generated pseudo code serves as the foundation for subsequent verification and refinement processes.
            \item \textbf{Verify Pseudo Code Logic:} In the second step, the generated pseudo code is subjected to a logic verification process. This step ensures that the pseudo code accurately reflects a viable solution to the problem by identifying and correcting any logical errors. The verified pseudo code provides a reliable blueprint for generating sample inputs and outputs.
            \item \textbf{Input and Output Sample:} Following the verification of pseudo code logic, the LLM takes sample input and output pairs from original question. These samples are used by LLM to better understand the question and play the important role for testing the correctness of the logical flow in pseudo code.
            \item \textbf{Verify Code Logic:} The focus here is on ensuring that the code from previous step behaves as expected when provided with the sample inputs, thereby producing the correct outputs. This verification process is essential for validating the logical coherence of the solution before final translating into real programming code.
            \item \textbf{Convert into Code:} After checked the logic of pseudo code, this steps focus on the implementation of translating pseudo code into real programming code.
            \item \textbf{Verify Code Logic:} The focus here is on ensuring that the translated code behaves as expected when provided with the pseudo code logic, sample inputs, thereby producing the correct outputs.
            \item \textbf{Input and Output Format:} The final step involves refining the code to adhere to the specified input and output format requirements. This step ensures that the code meets the problem's specifications in terms of structure and presentation. The output from this step is the finalized code, which is expected to solve the given problem accurately.
        \end{enumerate}
    
    \item \textbf{Prompt 6 (\texttt{all-in-one}):} This model includes all the prompt configurations from Prompt 5 (\texttt{multi}). However, instead of using a multi-step conversational process, all prompt settings are consolidated into one single prompt. The objective is to compare the efficacy of multi-step prompting versus an all-in-one approach.
\end{itemize}

By employing these diverse prompts, our study systematically investigates the influence of different prompt structures on the performance of LLMs, thereby providing insights into the optimal strategies for prompt engineering in the context of LLM-driven code assessment and guidance.


\subsection{Problem-Specific Prompts}
These are often unusual coding problems, commonly seen in coding competitions, complex optimization problem, or even number theory problems, where a universal guide prompt is ineffective. Detailed, problem-specific prompts are required to navigate the intricacies of such problems. For example, advanced dynamic programming challenges or problems involving intricate mathematical concepts typically require bespoke prompts that provide in-depth, step-by-step guidance tailored to the specific problem at hand. We will discuss this category in details in section \ref{section:usaco}.

\section{Experiments}
\label{section:experiments}

\subsection{Dataset}
\label{section:dataset}
Numerous sources and coding question banks (benchmarks) are available to assess the capabilities of AI code generation. The majority of these code questions are publicly accessible, having been extensively analyzed and discussed, thus serving as training data for most large language models (LLMs). Identifying a unique, diverse, and challenging coding dataset that has not been extensively used in training LLMs remains a significant challenge. In this study we used two data source as our dataset: 1) LeetCode 2) USACO.

\subsubsection{LeetCode Dataset}
We start with LeetCode \footnote{https://leetcode.com/} as our experinment dataset, due to its comprehensive range of well-defined and real-world-relevant programming problems. The structured nature of these problems enhances the AI's ability to process and generate solutions, providing a consistent benchmark for performance analysis. Furthermore, LeetCode holds significant educational value; it offers students practical applications of theoretical concepts, enhancing learning through exposure to diverse problem-solving techniques. For our experiments, we selected and compiled the first 100 questions from LeetCode (\texttt{lc100}), including both questions and test cases.\\

\noindent\fbox{\begin{minipage}{\columnwidth}
    \textbf{A Sample of LeetCode Question} \\
    
    \textbf{Problem Content:} Given an array of integers nums and an integer target, return indices of the two numbers such that they add up to target. You may assume that each input would have exactly one solution, and you may not use the same element twice. You can return the answer in any order. \\
    
    \textbf{Test Input:}
    nums    = [2, 7, 11, 15],
    target  = 9 \\
    \textbf{Test Output:} [0,1]
\end{minipage}}

\subsubsection{USACO Dataset}
Given that the LeetCode dataset has been widely published and discussed and extensively utilized for training by many current LLMs, \cite{brown2020language}, we aimed to identify a less commonly used dataset for our analysis. In this context, we have identified the United States of America Computing Olympiad (USACO) \footnote{https://usaco.org/}, an online computer programming competition that serves as a qualifier for the International Olympiad in Informatics in the USA, as a valuable data source. USACO provides a diverse set of coding problems, categorized by difficulty levels: Bronze, Silver, Gold, and Platinum. This dataset is particularly valuable because it is less commonly used as a training dataset by major LLMs, thereby offering a fresh and underutilized resource for evaluation. For our study, we selected relatively easy questions from the Bronze category, randomly picking 20 questions from different years spanning 2016 to 2023. This dataset is subsequently referred to as \texttt{usaco20}.

\noindent\fbox{\begin{minipage}{\columnwidth}
    \textbf{A Sample of USACO Question} \\
    
    \textbf{Problem Content:} Farmer John has lost his prize cow Bessie, and he needs to find her! Fortunately, there is only one long path running across the farm, and Farmer John knows that Bessie has to be at some location on this path. If we think of the path as a number line, then Farmer John is currently at position x and Bessie is currently at position y (unknown to Farmer John). If Farmer John only knew where Bessie was located, he could walk directly to her, traveling a distance of $|x-y|$. Unfortunately, it is dark outside and Farmer John can't see anything. The only way he can find Bessie is to walk back and forth until he eventually reaches her position. Trying to figure out the best strategy for walking back and forth in his search, Farmer John consults the computer science research literature and is somewhat amused to find that this exact problem has not only been studied by computer scientists in the past, but that it is actually called the "Lost Cow Problem" (this is actually true!). The recommended solution for Farmer John to find Bessie is to move to position $x+1$, then reverse direction and move to position $x-2$, then to position $x+4$, and so on, in a "zig zag" pattern, each step moving twice as far from his initial starting position as before. As he has read during his study of algorithms for solving the lost cow problem, this approach guarantees that he will at worst travel 9 times the direct distance $|x-y|$ between himself and Bessie before he finds her (this is also true, and the factor of 9 is actually the smallest such worst case guarantee any strategy can achieve).
    Farmer John is curious to verify this result. Given x and y, please compute the total distance he will travel according to the zig-zag search strategy above until he finds Bessie. \\
    \textbf{Testing Format:} \\INPUT FORMAT (file lostcow.in): The single line of input contains two distinct space-separated integers x and y. Both are in the range 0…1,000. \\OUTPUT FORMAT (file lostcow.out): Print one line of output, containing the distance Farmer John will travel to reach Bessie. \\
    \textbf{Testing Case:} \\
    SAMPLE INPUT:3 6 SAMPLE OUTPUT:9 
\end{minipage}}

\subsection{LLM Models}
\label{section:llm}
In this study, we recognize the vast and rapidly evolving landscape of LLMs, which includes numerous models of significance. Given the impracticality of listing all of them, we focus on the most popular and practically usable models, both open-source and closed-source. Our comparative analysis includes the following prominent LLMs: ChatGPT (gpt-4-turbo) \footnote{https://openai.com/research/gpt-4}, ChatGPT (gpt-4o) \footnote{https://openai.com/index/hello-gpt-4o/}, LLAMA (llama3-8b) \footnote{https://llama.meta.com/llama3/}, and Mistral (mixtral-8x7b) \footnote{https://mistral.ai}. Detailed configurations of these models are delineated in the Appendix under section \ref{section:prompt}.

\subsection{Evaluation Methods}
\label{section:evaluation}
For the LeetCode dataset, the website \footnote{https://leetcode.com/problemset/} provides all test cases and expected results for each question. Given that LeetCode questions are widely used in coding interviews, it is crucial to employ a comprehensive set of evaluation criteria that capture both functional correctness and code efficiency. Therefore, we utilize three primary indicators: pass rate (correctness), time spent, and Pylint score, each selected for their distinct contributions to a holistic evaluation.

\begin{enumerate}
    \item \textbf{Pass Rate}: The pass rate serves as a fundamental measure of correctness and reliability. It quantifies the proportion of test cases that a solution correctly handles, directly reflecting its ability to meet the problem's specifications under various scenarios. This metric is crucial as it directly correlates with the primary goal of any coding solution—its accuracy and functionality.
    \item \textbf{Time Spent}: This metric evaluates the efficiency of the problem-solving process. It encompasses the duration from initiating the coding of the solution to its successful execution and debugging. Monitoring time spent is essential for understanding the complexity and efficiency of the solution from a practical standpoint. In competitive programming, where time efficiency is as critical as correctness, this metric provides insights into the algorithm's performance under time constraints.
    \item \textbf{Pylint Score}: As a widely recognized tool for code analysis in Python, Pylint assesses code quality based on a set of coding standards and heuristics. The Pylint score is indicative of the maintainability, readability, and structural quality of code. High scores suggest adherence to Python coding conventions and best practices, which are pivotal for long-term code maintenance and clarity. By including this metric, the evaluation encompasses not only the functional aspects but also the quality of the coding practices employed.
\end{enumerate}

\begin{figure}
    \centering
    \includegraphics[width=0.8\linewidth]{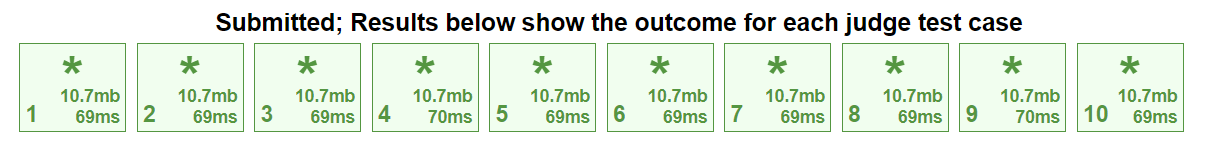}
    \caption{A screenshot of the USACO evaluation system displaying user submission results (all pass).}
    \label{fig:usaco_grading}
\end{figure}

For USACO dataset, their website provides a user submission evaluation system \footnote{https://usaco.guide/general/usaco-faq?lang=py}. All submitted code solutions are evaluated and scored against a set of predetermined test cases, considering not only correctness but also time and memory usage (Fig. \ref{fig:usaco_grading}). In this study, we use the USACO website to submit code generated by large language models (LLMs). A passing example is defined as code that successfully passes all test cases on their website.

At last, our experiment was conducted within a meticulously controlled and isolated virtual setting. By adopting this method, we guarantee the results' reliability and uniformity, offering a transparent and impartial evaluation of each LLM's effectiveness in addressing coding challenges.\footnote{We repeatedly tested using different high-quality code examples, and the results remained stable with no significant performance variations.}
\section{Result Analysis}
\label{section:result_analysis}

\subsection{LeetCode Results}
\label{section:leetcode}


\begin{table*}[ht]
\centering
\caption{Model Comparisons (Accuracy (\%), the higher, the better)}
\label{tab:result_acc}
\begin{tabular}{|r|r|r|r|r|r|r|}
\hline
\textbf{} & \textbf{base} & \textbf{example (1-shot)} & \textbf{dynamic example} & \textbf{guide} & \textbf{multi} & \textbf{all-in-one} \\ \hline
gpt-4        & 98\%          & \textbf{99\%} & \textbf{99\%} & \textbf{99\%} & \textbf{99\%} & 97\% \\ \hline
gpt-4o       & 97\%          & 98\%          & 88\%          & 97\%          & \textbf{100\%} & 96\% \\ \hline
llama3-8b    & \textbf{94\%} & 93\%          & 85\%          & 79\%          & 74\%          & 74\% \\ \hline
mixtral-8x7b & \textbf{75\%} & 66\%          & 66\%          & 75\%          & 67\%          & 74\% \\ \hline
\end{tabular}
\end{table*}

\begin{table*}[ht]
\centering
\caption{Model Comparison (Time Spent (ms), the lower, the better)}
\label{tab:result_time}
\begin{tabular}{|r|r|r|r|r|r|r|}
\hline
\textbf{} & \textbf{base} & \textbf{example (1-shot)} & \textbf{dynamic example} & \textbf{guide} & \textbf{multi} & \textbf{all-in-one} \\ \hline
gpt-4        & 4095 & 3988          & 3999 & 3994 & \textbf{3959} & 3984          \\ \hline
gpt-4o       & 4604 & 4431          & 4589 & 4430 & 4371          & \textbf{4129} \\ \hline
llama3-8b    & 4325 & 4173          & 4238 & 4216 & \textbf{4142} & 4196          \\ \hline
mixtral-8x7b & 4180 & \textbf{4097} & 4017 & 3954 & 3939          & 4032          \\ \hline
\end{tabular}
\end{table*}

\begin{table*}[ht]
\centering
\caption{Model Comparisons (Pylint Score, the higher, the better)}
\label{tab:result_pylint}
\begin{tabular}{|r|r|r|r|r|r|r|}
\hline
\textbf{} & \textbf{base} & \textbf{example (1-shot)} & \textbf{dynamic example} & \textbf{guide} & \textbf{multi} & \textbf{all-in-one} \\ \hline
gpt-4        & 9.59 & 9.58          & 9.63           & 9.56 & \textbf{9.66} & 9.38 \\ \hline
gpt-4o       & 9.34 & 9.49          & 9.25           & 9.00 & \textbf{9.62} & 9.52 \\ \hline
llama3-8b    & 9.18 & 9.14          & \textbf{10.00} & 8.64 & 8.41          & 7.24 \\ \hline
mixtral-8x7b & 9.05 & \textbf{9.17} & 8.27           & 8.04 & 8.27          & 7.74 \\ \hline
\end{tabular}
\end{table*}

In our experiments, we evaluated the performance of various models using different prompting strategies on easily solvable coding problems. The results, summarized in Table \ref{tab:result_acc}, indicate that GPT-4 and GPT-4o consistently outperform Llama3-8b and Mixtral-8x7b in terms of pass rate. Notably, GPT-4o achieved a perfect pass rate of 100\% with the "multi" prompt strategy, highlighting its adaptability and efficiency. GPT-4 also demonstrated high reliability with pass rates predominantly at 99\%, except for minor variations in the base scenario (98\%) and all-in-one prompt (97\%). Llama3-8b and Mixtral-8x7b showed lower adaptability, with Llama3-8b performing best in the base scenario (94\%) and declining in more complex prompts.

Regarding time efficiency, as shown in Table \ref{tab:result_time}, GPT-4 required the least time to execute across all prompts, with the shortest time recorded at 3959 milliseconds for the "multi" prompt. This suggests that GPT-4 is both effective and efficient. GPT-4o, while achieving the highest pass rate, exhibited slightly longer execution times, ranging from 4371 milliseconds ("multi") to 4604 milliseconds ("base"). Llama3-8b and Mixtral-8x7b generally required more time than GPT-4 but less than GPT-4o.

Pylint scores, reported in Table \ref{tab:result_pylint}, reveal that all models achieved high scores, indicating good adherence to coding standards. GPT-4 generally exhibited the highest Pylint scores, especially under the "multi" prompt, scoring 9.66. GPT-4o displayed variability in scores, with the highest being 9.62 ("multi") and the lowest 9.00 ("guide"). Llama3-8b and Mixtral-8x7b showed lower and more variable scores, with particularly low scores in the "all-in-one" and "guide" prompts, respectively. Notably, the "dynamic example" prompt resulted in a perfect score of 10.00 for Llama3-8b, though this was an outlier.

In summary, our findings indicate that while the base prompt and GPT-4 family models (GPT-4, GPT-4o) already exhibit exceptional performance, the multi-step prompt strategy offers limited improvement. This is primarily due to the widespread public discussion of questions like those on LeetCode, and the fact that the data used in training most LLMs, such as Common Crawl, already includes these questions and their solutions \cite{brown2020language}. Consequently, for simpler problems, more complex prompting strategies provide only incremental benefits, and using LLMs without additional prompts is sufficient.

\subsection{USACO Results}
\label{section:usaco}

\begin{table}[ht]
\centering
\caption{Results of USACO Dataset (Pass Example \& Accuracy)}
\label{tab:result_usaco}
\begin{tabular}{|p{2cm}|p{2cm}|p{3cm}|p{3.5cm}|p{2cm}|}
\hline
 &
  \textbf{\begin{tabular}[c]{@{}l@{}}Sovable\\ (base prompt)\end{tabular}} &
  \textbf{\begin{tabular}[c]{@{}l@{}}Solvable \\ (multi prompt)\end{tabular}} &
  \textbf{\begin{tabular}[c]{@{}l@{}}Solvable\footnote{The results for `Solvable (multi+spec prompt)' include additional specific prompts.} \\ (multi+spec prompt)\end{tabular}} &
  \textbf{\begin{tabular}[c]{@{}l@{}}Currently \\ Not Solvable\end{tabular}} \\ \hline
\begin{tabular}[c]{@{}l@{}}Question ID\\ (cpid)\end{tabular} &
  639, 737, 761, 766, 807, 939 &
  639, \textbf{641}, \textbf{735}, 737, \textbf{739}, \textbf{760}, 761, 766, 807, 939, \textbf{1228} &
  639, 641, 735, 737, 739, 760, 761, 766, \textbf{783}, \textbf{785}, \textbf{787}, 807, 939, 1228, \textbf{1323} &
  644, 738, 808, 1035, 1131 \\ \hline
Count \#         & 6    & 11   & 15   & 5    \\ \hline
\% (out of 20) & 30\% & 55\% & 75\% & 25\% \\ \hline
\end{tabular}
\end{table}

\footnotetext{Including solvable multi prompt and partial solvable spec prompts.}

We summarize all the pass examples in Table \ref{tab:result_usaco}. Given multi-step prompt (\texttt{multi}) and GPT-4o model demonstrated superior performance in our previous experiments on the LeetCode dataset, we compare three prompts (\texttt{base}, \texttt{multi}, and \texttt{spec}) with the GPT-4o model on the USACO dataset.

The base prompt configuration solved only 6 out of 20 problems (30\%). This indicates that LLMs struggle to generate valid code solutions without any form of prompt engineering. The low success rate suggests that minimal guidance in problem statements does not sufficiently leverage the capabilities of LLMs for effective problem-solving in computer programming.

In contrast, the multi prompt configuration showed significant improvement, solving 11 out of 20 problems (55\%). The Question IDs exclusive to the multi prompt (questions: 641, 735, 739, 760, 1228) are highlighted in bold in the table. This improvement underscores the advantages of multi-step conversational prompts. These prompts enhance the LLMs' contextual understanding, allow for iterative refinement, ensure logical coherence, and encourage user engagement and feedback. This method is particularly well-suited for complex problem solving as it incorporates thorough verification and validation processes, making it superior to the without-prompt (\texttt{base}) approach.

The multi+spec prompt configuration further extends the LLMs' capabilities, solving 15 out of 20 problems (75\%). This approach integrates specific question-related prompts (highlight in bold), such as analyzing the original code's time complexity and suggesting more efficient approaches (question 785), or listing all possible corner cases in a problem (question 783). Although these specific prompts cannot be generalized into a single prompt applicable to all questions, tailored prompts provide more detailed scenarios and considerations, thereby augmenting the LLMs' problem-solving abilities for complex tasks.

Despite these advancements, some problems remain unsolved by the LLMs, even with enhanced prompts (questions: 644, 738, 808, 1035, 1131). These challenging problems typically require complex logical reasoning, sequential decision-making, and optimization under constraints. They involve maintaining context across multiple stages, understanding intricate relationships between variables, and handling combinatorial tasks. For example, ensuring a farm remains fully connected after each barn closure or optimizing cow milking time based on pairings involves high-order planning and sophisticated problem-solving skills. These tasks often surpass the capabilities of current LLMs due to limitations in context retention, logical reasoning, and handling numerical and combinatorial complexities.

The results of our study indicate that prompt engineering significantly enhances the problem-solving capabilities of LLMs in the context of computer programming education. The multi-step prompts, in particular, demonstrate the potential to guide LLMs through complex problem-solving processes by providing a structured and iterative approach. We will discuss the usages of different strategies in prompt engineering in the next section.

\section{Discussion}
\label{section:discussion}

\subsection{LLM Model in Code Generation}
The results from our experiments indicate a clear recommendation for the use of the GPT-4 and its optimized variant GPT-4o as the preferred LLMs for educational purposes in computer programming instruction. The consistently high performance of these models across various prompting strategies highlights their adaptability, efficiency, and robustness. GPT-4o, in particular, demonstrated a perfect pass rate in the "multi" prompt strategy, suggesting its superior capability in understanding and generating accurate responses in complex problem-solving scenarios. The slightly higher execution times of GPT-4o compared to GPT-4 are a reasonable trade-off for its enhanced performance and adaptability. Therefore, we recommend GPT-4o as the primary model for educational settings, especially where complex problem-solving and iterative refinement processes are critical.

\subsection{Prompt Strategies Based on Educational Requirements}
Furthermore, to maximize the effectiveness of LLMs in different educational contexts, we propose tailored prompt strategies for three distinct educational requirements: regular knowledge and skills, competition preparation, and advanced problem-solving.

\begin{enumerate}
    \item \textbf{Knowledge and Skills}: For foundational learning and skill-building in computer programming, no-prompt engineering approaches suffice. This strategy involves presenting the LLM with questions that include problem content, constraints, and test cases. Since most fundamental computer science knowledge and exercises are already well-represented in LLM training datasets, this structured prompting method effectively aids in developing essential skills necessary for class exercise, basic algorithmic problem-solving and coding interviews. The method's systematic nature supports the learners in acquiring a solid foundation in programming concepts and techniques.
    \item \textbf{Competition Preparation} (e.g., USACO): In the context of preparing for programming competitions, such as USACO, the "Multi-Step Conversational Prompt" strategy is most effective. This approach allows for a dynamic interaction between the LLM and the student, where multiple turns of feedback and refinement are possible. The steps involved in generating pseudo code, verifying logic, and iteratively refining the solution are particularly beneficial for complex and competitive problems that require critical and innovative thinking. This strategy enhances the LLM’s contextual understanding and ensures logical coherence, making it ideal for high-stakes competitive scenarios.
    \item \textbf{Advanced Problem-Solving}: For tackling advanced and complex problems, the "Specific Prompt Engineering" strategy is recommended. This method provides detailed instructions and considerations specific to the problem at hand, such as treating the problem as an optimization task or considering extreme scenarios. The focused guidance helps in addressing sophisticated topics like algorithm design, optimization, and number theory. This strategy is essential for students who already have a strong grasp of basic concepts and are looking to deepen their understanding and tackle more intricate problems. The integration of tailored prompts, as seen in the "multi+spec" prompt configuration, further extends the LLMs’ capabilities, making it suitable for advanced educational objectives.
\end{enumerate}

\subsection{Future work}
As LLMs continue to advance, the use of multi-round prompts with various strategies remains vital in addressing complex problems. Future research will target several pivotal areas to augment LLM capabilities through prompt engineering. Initially, we will extend our testing of prompt engineering strategies on more challenging datasets, including more competition and number theory problems, to evaluate and refine our approaches across a broader spectrum of difficult issues. Furthermore, our research will emphasize improving context retention, enhancing logical reasoning, and better managing numerical and combinatorial complexities. Developing sophisticated prompt engineering techniques tailored to these challenging problem types is essential for optimizing the educational potential of LLMs. By systematically categorizing and testing various prompt engineering strategies, we aim to establish a robust framework incorporating retrieval-augmented generation (RAG), multi-agents system, and Plan-and-Execute tools for educational implementation. This framework will provide educators with comprehensive guidelines to optimize LLM-based learning experiences, ultimately enhancing educational outcomes in computer programming instruction.

\section{Conclusion}
\label{section:conclusion}
This paper explores the potential of prompt engineering in large language models (LLMs) to enhance educational outcomes in computer programming instruction. Our research is driven by three key questions: the systematic categorization of prompt engineering strategies tailored to educational requirements, the empowerment of LLMs to solve complex problems, and the establishment of a robust framework for testing and implementing these strategies.

Our findings indicate that the GPT-4 and GPT-4o models outperform other LLMs such as Llama3-8b and Mixtral-8x7b in terms of pass rates, execution times, and adherence to coding standards. The GPT-4o model, in particular, demonstrated a successfully pass rate with the "multi" prompt strategy, highlighting its superior adaptability and efficiency. These results lead us to recommend GPT-4o as the preferred model for educational purposes in computer programming.

We propose tailored prompt strategies based on educational requirements. For foundational learning and skill-building, such as LeetCode, ask question directly without prompt engineering is suffice, providing structured guidance that helps students grasp essential concepts and techniques. For competition preparation, such as USACO, the "Multi-Step Conversational Prompt" strategy proves beneficial, facilitating dynamic interaction and iterative refinement that enhance contextual understanding and problem-solving skills. For advanced problem-solving, the "Specific Prompt Engineering" strategy is ideal, offering detailed instructions that address complex topics like algorithm design and optimization.

Our study also highlights the significant role of prompt engineering in maximizing the potential of LLMs in educational contexts. By categorizing and testing various strategies, we have established a robust framework for their implementation, providing educators with comprehensive guidelines to optimize LLM-based learning experiences. Despite the advancements, certain complex problems remain challenging for current LLMs, suggesting the need for further research to enhance context retention, logical reasoning, and handling of numerical and combinatorial complexities.

In conclusion, prompt engineering significantly enhances the capabilities of LLMs in computer programming education. The tailored strategies we propose align with specific educational objectives, from foundational learning to advanced problem-solving. The superior performance of GPT-4 and GPT-4o confirms their suitability for a wide range of educational applications. By adopting and refining these strategies, educators can significantly improve educational outcomes, providing students with a more effective and personalized learning experience in computer programming.

\bibliographystyle{unsrtnat}

\newpage
\appendix
\section{Prompts in This Research}
\label{section:prompt}
Within the domain of LLMs, such as the entity engaged in the current interaction, the terminologies \texttt{system prompts} and \texttt{user prompts} delineate distinct categories of input stimuli instrumental in directing the model's output generation.

\begin{itemize}
    \item \textbf{System Prompt}: The system prompt is used to define the persona that LLM will play.  In our application, it is part of the system's design to help guide the LLM in generating responses or performing code generation to ensure it plays as a sophisticated programmer professional or mentor. 
    \item \textbf{User prompts}: They are the prompts specifying the questions, commands, or statements that users input into the system to seek a response from the LLM (played as a programming tutor). In essence, this encompasses the textual or verbal input furnished by the user to the model.
\end{itemize}

The following figures presents illustrative examples of prompts utilized within the scope of our manuscript.

\begin{figure}[ht]
    \centering
    \includegraphics[width=1.0\linewidth]{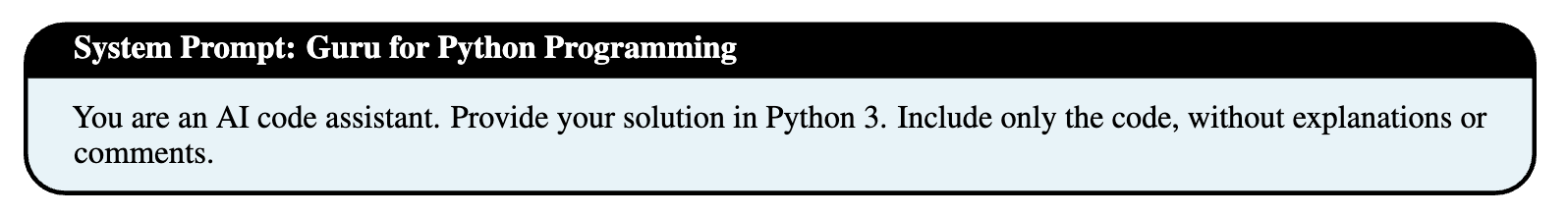}
    \caption{System Prompt}
    \label{fig:prompt_1}
\end{figure}

\begin{figure}[ht]
    \centering
    \includegraphics[width=1.0\linewidth]{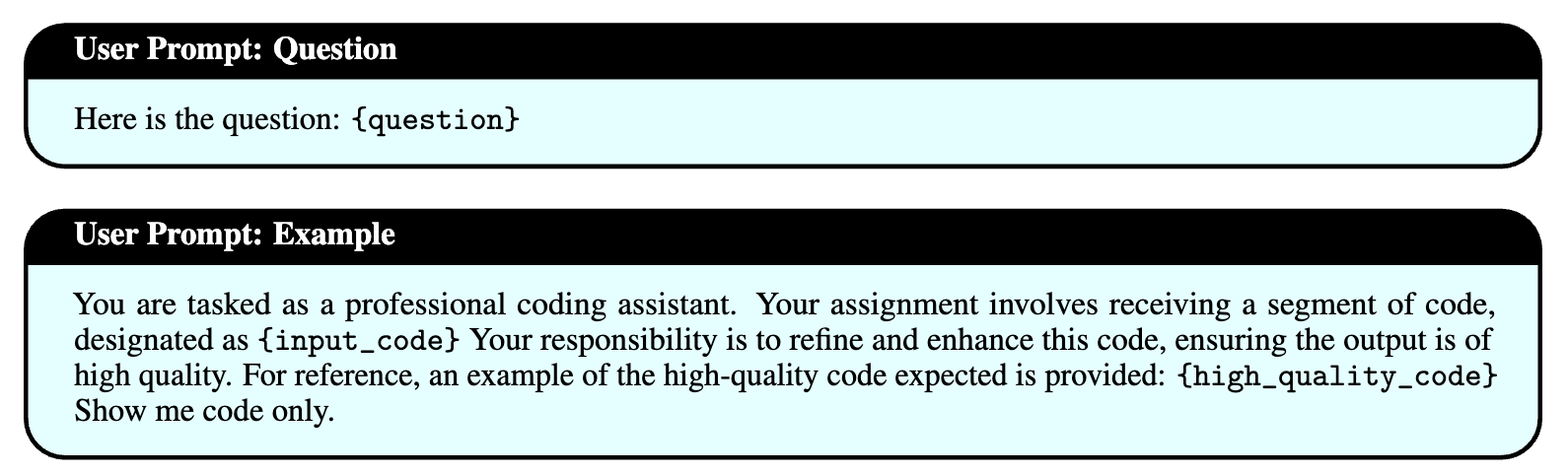}
    \caption{User Prompt (Base and Example)}
    \label{fig:prompt_2}
\end{figure}

\begin{figure}[ht]
    \centering
    \includegraphics[width=1.0\linewidth]{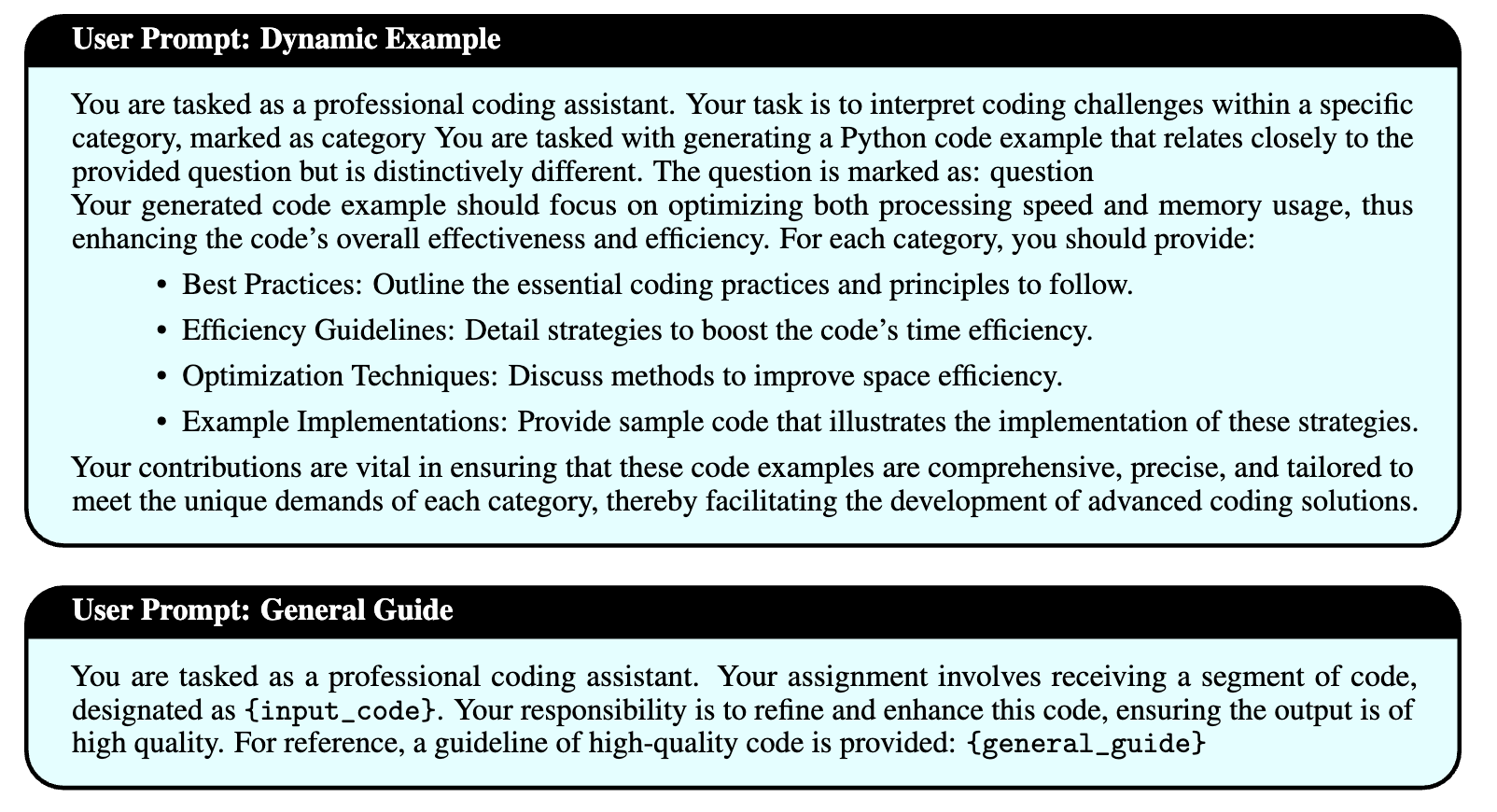}
    \caption{User Prompt (Dynamic Example, General Guide)}
    \label{fig:prompt_3}
\end{figure}

\begin{figure}[ht]
    \centering
    \includegraphics[width=1.0\linewidth]{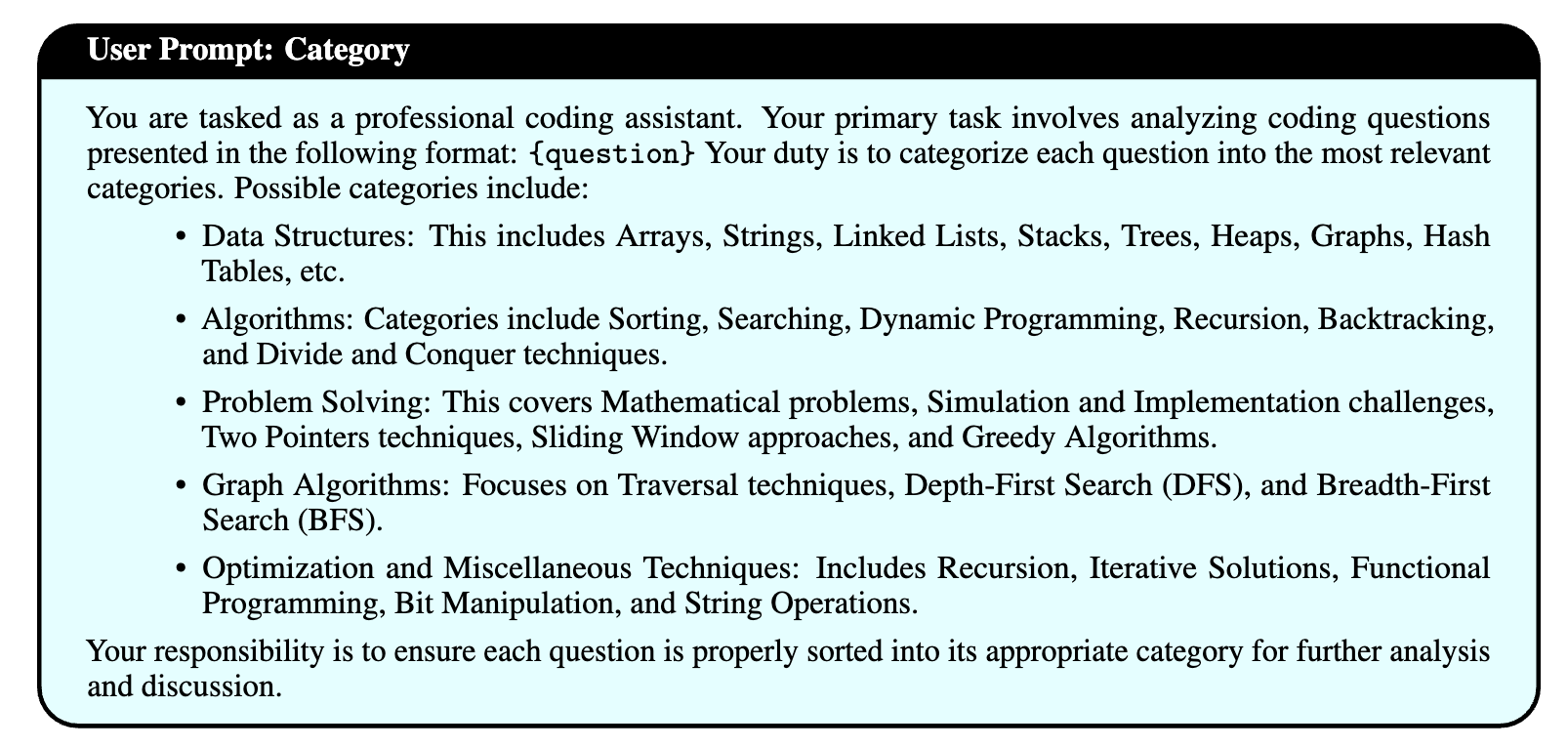}
    \caption{User Prompt (Category Question)}
    \label{fig:prompt_4}
\end{figure}

\begin{figure}[ht]
    \centering
    \includegraphics[width=1.0\linewidth]{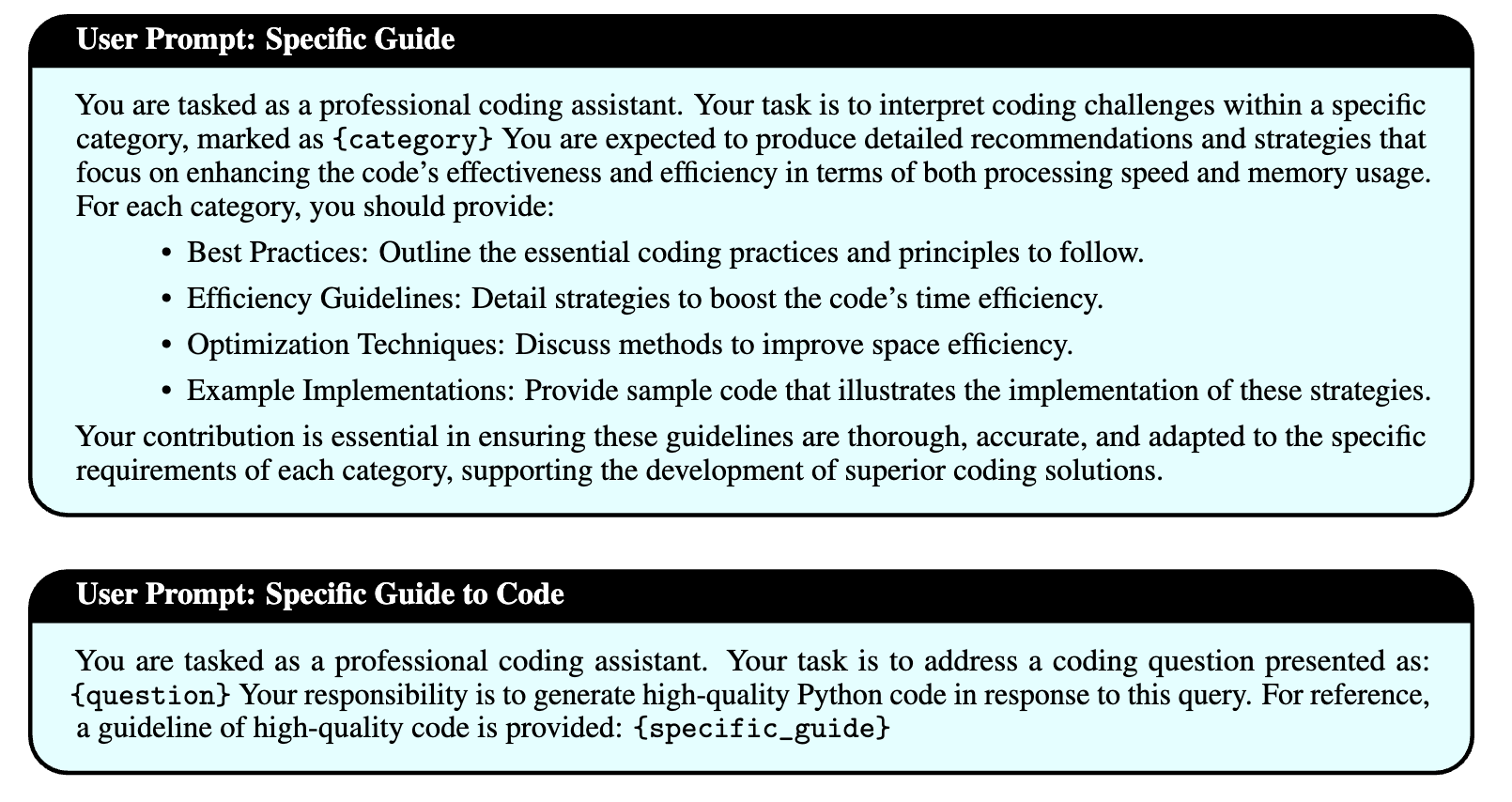}
    \caption{User Prompt (Category and Guide)}
    \label{fig:prompt_5}
\end{figure}

\begin{figure}[ht]
    \centering
    \includegraphics[width=1.0\linewidth]{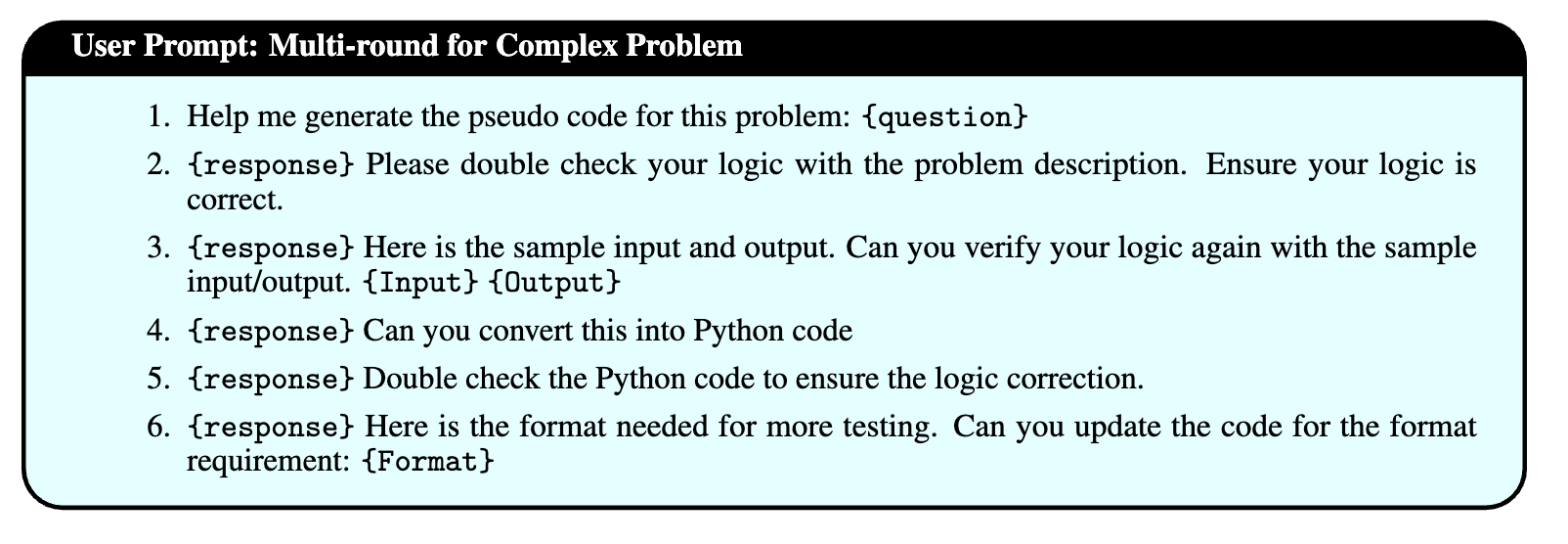}
    \caption{User Prompt (Multi-round Guide)}
    \label{fig:prompt_6}
\end{figure}
\section{General Guidelines for High-Quality Python Code}
\label{section:general_guide_content}
\begin{enumerate}
    \item \textbf{Use of Standard Libraries}:
    The code effectively utilizes Python's standard libraries such as re (for regular expressions), heapq (for heap queue algorithms), bisect (for array bisection algorithms), and math.\\
    Best Practice: Always prefer standard libraries for common algorithms. Do not import unnecessary libraries.
    \item \textbf{Use Proper Data Structures}:
    Fundamental building blocks for linked lists and binary trees. These structures are essential for representing hierarchical or sequential data.
    Stack: Implements a basic data structure, like Node, ListNode, TreeNode: a stack, a queue, a linked list, or a binary tree.\\
    Best Practice: Create and choose the right data structure based on the problem requirements for efficiency in both time and space complexity.
    \item \textbf{Itertools Usage}:
    The use of itertools for permutations, combinations, and cartesian product is a sign of advanced Python knowledge. These functions are crucial for solving combinatorial problems.\\
    Best Practice: Leverage itertools to write cleaner and more efficient looping constructs.
    \item \textbf{Heap Queue (heapq)}:
    The heapq module is used for implementing priority queues. This is vital in algorithms where you need to repeatedly access the smallest or largest element.\\
    Best Practice: Use heapq for priority queue implementation if necessary instead of a sorted list for better performance.
    \item \textbf{Functional Programming}:
    Usage of functools like reduce, cache, and lru\_cache indicates a functional approach to problem-solving. This is efficient for operations that benefit from memoization or reducing a list.\\
    Best Practice: Employ functional programming concepts where applicable to make the code more concise and readable.
    \item \textbf{Math and Random Modules}:
    The inclusion of math operations and random for random number generation is suitable for problems involving math computations and stochastic processes.\\
    Best Practice: Understand and utilize the vast array of functions provided by these modules to simplify complex calculations.
    \item \textbf{Efficiency and Optimization}:
    The use of bisect for binary searches and heapq for efficient element access in priority queues shows an understanding of algorithmic efficiency.\\
    Best Practice: Always consider time and space complexity; optimize code where necessary but avoid premature optimization.
    \item \textbf{Number Theory}:
    When solving problems related to number theory, it is essential to leverage mathematical properties and efficient algorithms. Utilize functions from the math module for operations like finding greatest common divisors (gcd), least common multiples (lcm), and performing modular arithmetic.\\
    Best Practice: Use built-in functions and algorithms for common number theory operations. Optimize prime-related computations using the Sieve of Eratosthenes for prime generation and employ efficient algorithms for primality testing and factorization to handle large numbers effectively.
\end{enumerate}
\section{Hyper-parameters in LLMs}




\begin{table}[ht]
\centering
\caption{Hyper-Parameter of LLMs}
\label{tab:hyper_parameters}
\begin{tabular}{|c|c|}
\hline
\textbf{Parameter}            & \textbf{Models}                                                                                                             \\ \hline
Engine               & \begin{tabular}[c]{@{}c@{}}gpt-4-turbo, gpt-4o,\\ \\ meta-llama-3-8b-instruct,\\ \\ open-mixtral-8x7b\end{tabular} \\ \hline
Max Token            & 4096                                                                                                               \\ \hline
n (responses)        & 1                                                                                                                  \\ \hline
Temperature          & 0.7                                                                                                                \\ \hline
Top P                & 1                                                                                                                  \\ \hline
Frequency Penalty    & 0                                                                                                                  \\ \hline
Presentation Penalty & 0.6                                                                                                                \\ \hline
\end{tabular}
\end{table}

In this research paper, we choose and compare three popular LLMs models: \texttt{gpt-4-turbo}, \texttt{gpt-4o},  \texttt{meta-llama-3-8b-instruct} and \texttt{open-mixtral-8x7b}. The details configurations can be found at Table \ref{tab:hyper_parameters}. We set the maximum token limit to 4096, allowing for extensive and detailed responses. Our experiments were conducted with a single response output (\(n=1\)), ensuring focused and specific replies and each LeetCode question will only generate one code solution. To balance creativity with relevance, we use the default temperature 0.7, a moderate setting that encourages a mix of predictable and innovative responses. The \texttt{top\_p} parameter was set to 1, enabling the model to consider the full range of possible next words. We applied a frequency penalty of 0.0 to prevent repetition and a presence penalty of 0.6, encouraging the model to introduce new concepts and ideas throughout the conversation. This configuration was pivotal in achieving the desired balance between coherence, relevance, and novelty in the model's responses.

\end{document}